\documentclass[a4paper]{article}
\usepackage{INTERSPEECH2019}

\usepackage{breqn}
\usepackage{graphicx}
\usepackage{subcaption}
\usepackage[font=footnotesize,labelfont=bf]{caption}
\usepackage[pagebackref=true,breaklinks=true,letterpaper=true,colorlinks,bookmarks=false]{hyperref}
\usepackage{booktabs}
\usepackage{diagbox} 
\usepackage{amssymb}
\usepackage{pifont}
\usepackage{textcomp}
\usepackage{xcolor}
\usepackage{nicefrac}
\definecolor{lred}{HTML}{E99A93}
\definecolor{lyellow}{HTML}{FFE591}
\definecolor{lblue}{HTML}{A3C3F3}

\newcommand{\cmark}{\ding{51}}%
\newcommand{\xmark}{\ding{55}}%

\def\newpara{\vspace{2pt}}

\title{My lips are concealed: Audio-visual speech enhancement through obstructions }
\name{Triantafyllos Afouras$^1$, Joon Son Chung$^{1,2}$, Andrew Zisserman$^1$}
\address{
  $^1$Visual Geometry Group, Department of Engineering Science, University of Oxford\\
  $^2$Naver Corporation}
\email{\{afourast,joon,az\}@robots.ox.ac.uk}

\begin{document}

\maketitle
\begin{abstract}

Our objective is an audio-visual model for separating a single speaker
from a mixture of sounds such as other speakers and background
noise. Moreover, we wish to hear the speaker even when the visual cues
are temporarily absent due to occlusion.

To this end we introduce a deep audio-visual speech enhancement
network that is able to separate a speaker's voice by conditioning on
both the speaker's lip movements and/or a representation of their
voice.  The voice representation can be obtained by either (i)
enrollment, or (ii) by self-enrollment -- learning the representation
on-the-fly given sufficient unobstructed visual input.  The model is
trained by blending audios, and by introducing artificial
occlusions around the mouth region that prevent the visual modality
from dominating.

The method is speaker-independent, and we demonstrate it on real examples of speakers unheard
(and unseen) during training. The method also improves
over previous models in particular for cases of occlusion in the visual modality.

\end{abstract}

\section{Introduction}

  While there has been great progress in the field of automatic speech recognition (ASR) in recent years, 
  some key challenges remain, particularly the understanding of speech in very noisy environments or in
  cases where multiple people speak simultaneously.  
  In this direction, isolating voices in multi-speaker scenarios, increasing
  the signal-to-noise ratio in noisy audio, or combinations of both are all important tasks.

  Until very recently, works in this area have only used the audio modality for the task. 
  However, recent works have shown that the use of video can aid tremendously in solving 
  the problem~\cite{Afouras18, Ephrat18, Owens18}.

\begin{figure}[t]
  \includegraphics[width=\linewidth]{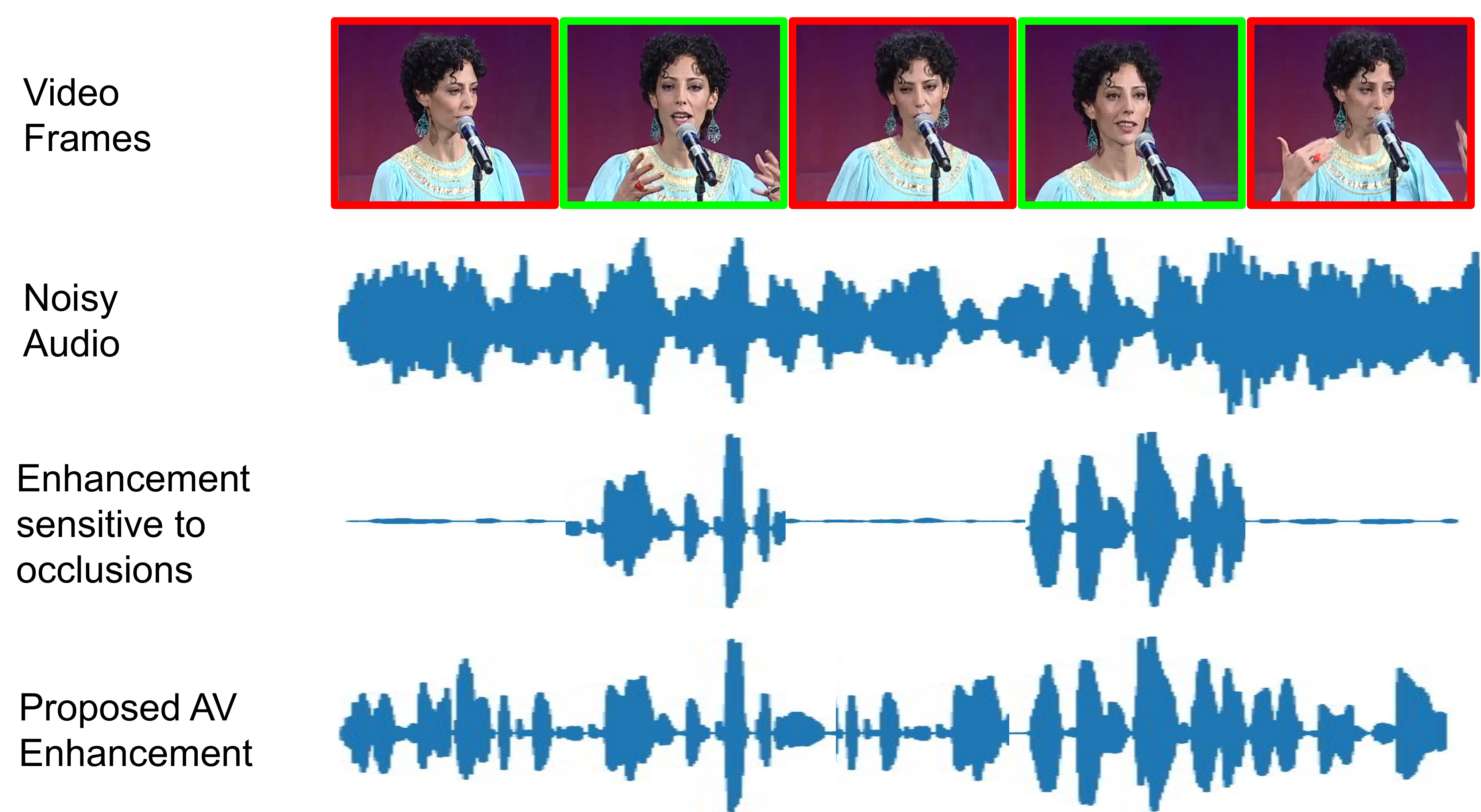}
  \vspace{-10pt}
  \caption{An audio-visual speech enhancement model may fail when the lip
  region is occluded by e.g. a microphone. In such cases the
  input audio is often entirely filtered out and the result is silent output over the occluded frames. The aim of our
  method is to be robust to this kind of occlusions. }
  \label{fig:teaser}
  \vspace{-15pt}
\end{figure}

  These audio-visual models have demonstrated impressive results, but given their
  dependence on the visual input, they may fail when
  the mouth area is occluded by the speaker's hands, 
  a microphone (e.g. Fig.~\ref{fig:teaser}), or if the speaker turns their head away.
  Contemporaneously, it has been shown that an embedding of the speaker's voice can guide
  the separation of simultaneous speech~\cite{wang2018voicefilter}.

In this paper we  propose combining the two approaches, i.e.\ conditioning on both the video input containing the
  speaker's lip movement and an embedding of their voice, in order to make the audio-visual models 
robust to occlusions.
  Our assumption is that the video provides invaluable discriminative information when present,
  while the speaker embedding can help the  model when the video is absent due to occlusions.
In the simplest case, the voice embedding can be obtained from pre-enrolled audio. 

While it is possible to separate simultaneous speakers using only the
audio~\cite{hershey2016deep,yu2016}, the permutation issue in the
time-domain remains an unsolved problem. With our approach, even
partially occluded video can provide  information on the
voice characteristics of the speaker and resolve the ambiguity of
assigning the separated voice to the speaker.

 We make the following contributions:
 (i) we show how speaker embedding and visual cues can be combined to separate a single
speaker from a mixture of voices
despite the visual stream (the lips) being occluded;
(ii) we propose a neural network model that can operate with
video only, enrollment data only, or both; and 
(iii) we introduce a recurrent model that can bootstrap the computation of the speaker embedding under
 temporary occlusions, without requiring a prior speaker embedding. We term this {\em self-enrollment}.


\subsection{Related Work}

\newpara\noindent{\textbf{Audio-only enhancement and separation.}}
  Various methods have been proposed to isolate multi-talker simultaneous speech,
  the majority of which only use monaural audio,
  {\em e.g.}~\cite{reddy2007soft,jin2009supervised,radfar2007single,makino2007blind,wang2017supervised}.
A number of recent works have addressed the permutation problem to separate unseen speakers.
 Deep clustering ~\cite{hershey2016deep} uses embeddings trained to yield a low-rank approximation to an ideal pairwise affinity matrix, whilst Yu~{\em et al.}
 employ a permutation invariant-loss \cite{yu2016}.

  \newpara\noindent{\textbf{Audio-visual speech enhancement.}}
  Prior to the advent of deep learning, numerous works have been
  developed for audio-visual speech
  enhancement~\cite{khan2013speaker,wang2005video,girin2001audio,deligne2002audio,hershey2002audio,Rivet14}.
  Several recent methods have used a deep learning framework for the same task --
  most notably \cite{Gabbay17a,Gabbay17b,Hou2017}.
  However, these methods are limited in that they are only demonstrated under constrained
  conditions (e.g.\  the utterances consist  of a fixed set of phrases),  or for a 
  small number of known speakers.
  Our previous work~\cite{Afouras18} proposed a deep audio-visual speech
  enhancement network that is able to separate a speaker's voice given
  lip regions in the corresponding video, by predicting both the magnitude and the
  phase of the target signal.
  Ephrat {\it et al.}~\cite{Ephrat18} designed a network that conditions on the video input of all
  the source speakers and outputs complex masks, thus also enhancing both magnitude and phase.
  Owens and Efros~\cite{Owens18} train a network on audio-visual synchronization and use the learned
  features for speaker separation. These last works demonstrate general results in-the-wild case.

  \newpara\noindent{\textbf{Enhancement by conditioning on voice only.}} Wang {\it et al.}~\cite{wang2018voicefilter} develop a
  method that separates voices conditioned on pre-learned speaker embeddings, showing that voice
  characteristics alone can be enough to determine the separation.
  This however relies on a pretrained model and does not use video.

  We propose to combine the two ideas: using both visual input and voice embeddings from the target
  speaker; our method partially builds on ~\cite{Afouras18, Ephrat18, wang2018voicefilter}.

\section{Method} \label{sec:method}

\begin{figure}[t]
      \centering
      \includegraphics[width=\columnwidth]{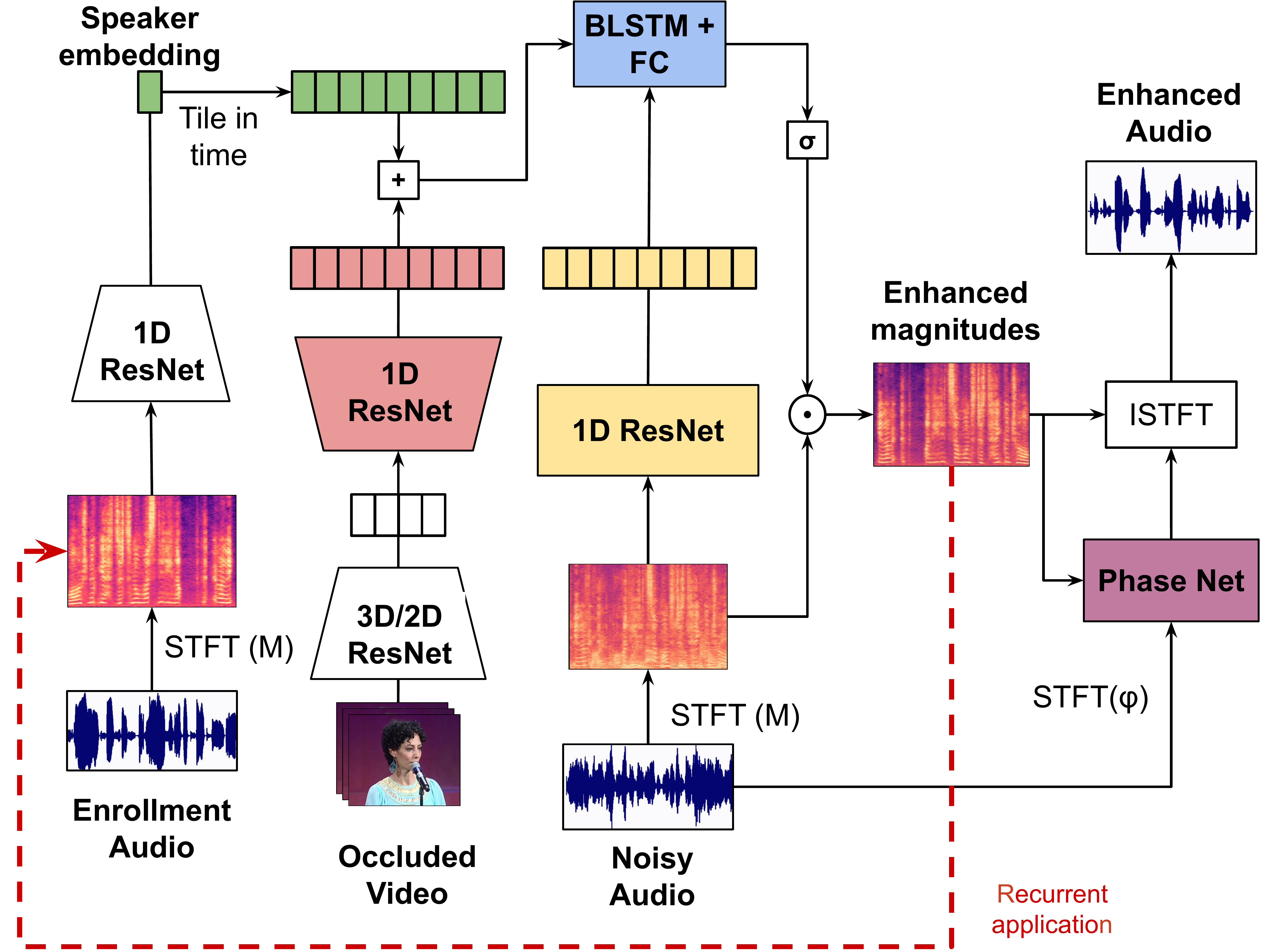}
      \vspace{-10pt}
      \caption{The architecture of the audio-visual speech enhancement network: 
       There are 2 audio streams. The one processes the incoming noisy
       audio, while the other takes as input an enrollment audio sample and creates a speaker embedding that captures the speaker's voice
       characteristics. A visual stream extracts frame-wise representations from the input video.
       The visual, speaker and audio embeddings are combined and fed into the
       BLSTM which outputs a multiplicative mask that filters the noisy spectrograms. When no
       enrollment audio is provided, the enhanced magnitudes (created by a video-only pass) can be
       used as the input to the speaker embedding network. }
      \label{fig:architecture}
      \vspace{-15pt}
\end{figure}

This section describes the architecture of the audio-visual speech enhancement network, which is
given in Figure~\ref{fig:architecture}.
The network receives three inputs: (i) the noisy audio to be enhanced; (ii) the corresponding video
frames; (iii) a reference audio containing speech from the target speaker.
We summarize the principal modules below.
Details of the architecture are provided in Table \ref{table:twostream_vert}.

\begin{table}[t]
\setlength{\tabcolsep}{2pt}
\setlength\extrarowheight{0.5pt}
\footnotesize
\begin{subtable}[t]{0.49\linewidth}\centering
\colorbox{lred}{%
\begin{tabular}[t]{  l r r r r r }
 \toprule
 Layer & \# filters & K & S   &  P & Out  \\  
 \midrule
 fc0    & 1536 &1 & 1  &  1 & $\nicefrac{T}{4}$  \\  
 conv1 & 1536 & 5 & 1  &  2 & $\nicefrac{T}{4}$  \\  
 conv2 & 1536 & 5 & 1  &  2 & $\nicefrac{T}{4}$  \\  
 conv3 & 1536 & 5 & \textonehalf &  2 & $\nicefrac{T}{2}$  \\  
 conv4 & 1536 & 5 & 1  &  2 & $\nicefrac{T}{2}$  \\  
 conv5 & 1536 & 5 & 1  &  2 & $\nicefrac{T}{2}$  \\  
 conv6 & 1536 & 5 & 1  &  2 & $\nicefrac{T}{2}$  \\  
 conv7 & 1536 & 5 & \textonehalf  &  2 & T  \\  
 conv8 & 1536 & 5 & 1  &  2 & T  \\  
 conv9 & 1536 & 5 & 1  &  2 & T  \\  
 fc10  & 256 & 1 & 1  &  1 & T  \\  
 \bottomrule
\end{tabular}
}
\caption{Video Stream}
\end{subtable}
\begin{subtable}[t]{0.49\linewidth}\centering
\colorbox{lyellow}{%
\begin{tabular}[t]{  l r r r r r }
 \toprule
 Layer & \# filters & K & S   &  P & Out  \\  
 \midrule
 fc0   & 1536 & 1 & 1  &  1 & T  \\  
 conv1 & 1536 & 5 & 1  &  2  & T  \\  
 conv2 & 1536 & 5 & 1  &  2  & T  \\  
 conv3 & 1536 & 5 & 1  &  2  & T  \\  
 conv4 & 1536 & 5 & 1  &  2  & T  \\  
 conv5 & 1536 & 5 & 1  &  2  & T  \\  
 fc6   & 256 & 1 & 1  &  1 & T  \\  
 \bottomrule
\end{tabular}
}%
\caption{Noisy Audio Stream}
\colorbox{lblue}{%
\begin{tabular}{  l r r  }
 \toprule
 Layer & \# filters & Out  \\  
 \midrule
 BLSTM   & 400 & T  \\  
 fc1     & 600 & T  \\  
 fc2     & 600 & T  \\  
 fc\_mask     & F & T  \\  
 \bottomrule
\end{tabular}
}%
\caption{AV Fusion}
\end{subtable}

\normalsize
\caption{Architecture details.
{\bf a)} The 1D ResNet that processes the video features.
{\bf b)} The 1D ResNet that processes the noisy audio spectrogram.
  {\bf c)} The BLSTM and FC layers that perform the modality fusion.
  Notation: {\bf K:}  Kernel width; {\bf S:} Stride -- fractional strides denote transposed convolutions;
  {\bf P:} Padding; {\bf Out:} Temporal dimension of the layer's output.
  The non-transposed convolution layers are all depth-wise separable.
  Batch Normalization, ReLU activation and a shortcut connection are added after every
  convolutional layer. }
\label{table:twostream_vert}
\vspace{-30pt}

\end{table}


  \newpara\noindent{\textbf{Video representation.}}
Input to the network is pre-cropped image frames, such as the face crops found in the LRS
datasets~\cite{Afouras19,Afouras18d}.
Visual features are extracted from the sequence of image frames using a spatiotemporal residual
network described in ~\cite{Stafylakis17}.
The network contains a 3D convolution layer, followed by a common 18-layer 2D ResNet~\cite{He16}.
For every video frame it outputs a compact $512$ dimensional feature vector.

\newpara\noindent{\textbf{Audio representation.}}
As acoustic features, we use the magnitude and phase spectrograms extracted from the audio waveforms
using a Short Time Fourier Transform (STFT)
with a 25ms window length and a 10ms hop length at a sample rate of 16kHz. This results in
spectrograms with a time dimension four times the number of corresponding video frames. We 
use $T/4$ and $T$ to denote the number of video frames and corresponding time resolution of the
spectrograms respectively.

\newpara\noindent{\textbf{Speaker embedding network.}}
For embedding a reference audio clip into a compact speaker representation, we use the method of 
Xie {\it et al.}~\cite{Xie19a}. To reduce the number of computations,
we replace all 2D spatial convolutions with 1D temporal ones
which regard the frequency bins as channels
and pre-train the modified architecture on the VoxCeleb2~\cite{Chung18a} dataset following
\cite{Xie19a}.

\newpara\noindent{\textbf{Modality combination.}}
As shown in Figure~\ref{fig:architecture}, the noisy magnitude spectrograms are encoded into
audio feature vectors through a shallow temporal ResNet.
The video features are upsampled through a network containing two transposed
convolution layers to match the temporal dimension of the spectrograms ($4T$).
The speaker embedding extracted from the reference audio is tiled temporally and added to
the resulting video embeddings to form the conditioning vector used for the enhancement.  
This vector is then fed along with the noisy audio embedding 
into a one-layer bidirectional LSTM, followed by two fully connected layers.
The output has spectrogram dimensions and
is passed through a sigmoid activation to produce the enhancement mask.


\newpara\noindent{\textbf{Phase sub-network.}}
In order to adjust the noisy phases to the enhanced magnitudes, we use the phase network
of~\cite{Afouras18} without any changes.

\newpara\noindent{\textbf{Self-enrollment.}}
For self-enrollment, the magnitude network is run twice:
on the first pass, no speaker embedding is added to the visual one. 
The magnitudes that are output then serve as input to the speaker embedding network, as indicated by the
red feedback arrow, and the network is run for a second time, with speaker embeddings this time. 

\noindent
We minimize the learning objective~\cite{Afouras18}:
\begin{equation}
 \mathcal{L} = ||\hat{M} - M^*||_{1} \ - \  \ \frac{1}{TF}  \sum_{t,f} M_{tf}^* 
  <\hat{\Phi}_{tf}, \Phi^*_{tf}> \nonumber
\end{equation}
where $\hat{M}$, $\hat{\Phi}$ and
$M^*$, $\Phi^*$ are the predicted and ground truth magnitude and phase
spectrograms respectively, and $T$ and $F$ their time and frequency resolutions.

\section{Experimental Setup}

\newpara\noindent{\textbf{Datasets.}}
The network is trained on the MV-LRS~\cite{Chung17a}, LRS2~\cite{Afouras19}, and LRS3~\cite{Afouras18d} datasets,
and tested on LRS3.
MV-LRS and LRS2 contain material from British television broadcasts, while LRS3 was created from
videos of TED talks. The speakers appearing in
LRS3 are to the best of our knowledge not seen in either of the other two datasets.
The datasets share the same format and pipeline including the
face detection step, therefore no pre-processing is required in order to utilise them together for training.
We remove from the LRS3 training set the few speakers that also appear in the test set, so that
there is no overlap of identities between the two.
Hence, the test set contains only speakers unseen and unheard during training and is suitable for a
speaker-agnostic evaluation of our methods.
Moreover, since the test set of LRS3 contains relatively short sentences,
for testing we extract some longer sub-sequences from the original material used to make the LRS3
test set.
We only use samples from speakers that appear in at least 2 different videos (TED talks), to
enable enrollment with audio recorded in a different setting than the target one.   
These extra videos, along with the added noise and occlusions have been made publicly available on
the project website.

\newpara\noindent{\textbf{Synthetic data.}}
We generate synthetic examples similarly to other works \cite{Afouras18, Ephrat18, wang2018voicefilter}
by first sampling one reference audio-visual utterance from the training dataset and then
mixing its audio with interfering audio signals. We consider two scenarios: 2 speakers and 3 speakers,
where one and two interfering voices are added to the target signal respectively.

\newpara\noindent{\textbf{Enrollment.}}
During training we do not know the identities of the speakers.  
Therefore, we obtain the enrollment signal from the same video but a different, non-overlapping time segment.
This effectively reduces the amount of data we can use as we need to
discard shorter videos (e.g. if we use 3 seconds, we can only use videos at least 6 seconds long). We
use this method for training on datasets where the speaker identities are not known.

During evaluation we experiment with two enrollment methods:
(i) {\it pre-enrollment --} we sample an enrollment segment from a video of the same speaker that is different from the one
used to create the target sample (we do have identity labels for the test set); 
(ii) {\it self-enrollment --} we obtain the enrollment audio with a pass through our network that does not use a speaker
embedding, as
explained in Section~\ref{sec:method}.

\begin{figure}[t]

  \begin{subfigure}{0.49\columnwidth}
      \includegraphics[width=0.49\linewidth]{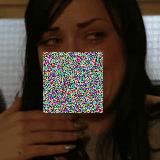}
      \includegraphics[width=0.49\linewidth]{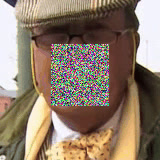}
      \caption{Training samples}
      \label{fig:train_samples}
  \end{subfigure}
  \begin{subfigure}{0.49\columnwidth}
      \includegraphics[width=0.49\linewidth]{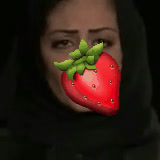}
      \includegraphics[width=0.49\linewidth]{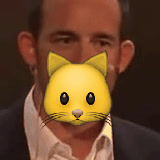}
      \caption{Test samples}
      \label{fig:test_samples}
  \end{subfigure}
      \label{fig:occlusions}
  \vspace{-5pt}
  \caption{Example frames of occluded videos used during training end evaluation.}
  \vspace{-20pt}
\end{figure}

\newpara\noindent{\textbf{Occlusions.}}
For training, we artificially add occlusions to the video frames in the form of random patches as shown in Figure~\ref{fig:train_samples}.
We randomly occlude sub-sequences of 15 to 25 contiguous frames, maintaining the clear-to-occluded
frames ratio at 1:3.
This is more realistic than simply zeroing out the incoming visual frames, as occluded
video frames still produce valid feature vectors. 
For evaluation however, instead of random patches, we place jittering emojis on the videos
as shown in Figure~\ref{fig:test_samples}. This type of visual noise has not been
seen during training.
The emojis are used to occlude the video from the start  and the end, while the middle of the utterance is kept clear.

\newpara\noindent{\textbf{Training.}}
The spatio-temporal visual front-end is pre-trained on a word-level lip reading task
\cite{Stafylakis17}.
We then freeze the front-end and pre-compute the visual features. 
The features are extracted on a version of the videos where we have added random occlusions.

Training is conducted in four phases.
We first pre-train the magnitude subnetwork only with speaker embedding inputs. For this we first use
mixtures of two and then three speakers.
Second, the visual modality is added and the magnitude network is trained on the saved visual features
for the three simultaneous speakers scenario; third the magnitude network is frozen and the phase network is trained; finally the whole
network is trained end-to-end.


\section{Experiments}

\subsection{Evaluation protocol}
To evaluate the performance of our model we use the Signal to Distortion Ration
(SDR)~\cite{Fevotte05}, a common metric
expressing the ratio between the energy of the target signal and of the errors contained in the enhanced output.
Furthermore to assess the intelligibility of the output, we use the Google Cloud ASR system --
we compute the Word Error Rate (WER) between the prediction of the ASR system on the enhanced audio
and the ground truth transcriptions of the utterances contained in the segments used for
evaluation.
We evaluate on fixed length video segments of 8 seconds (200 frames).
Some additional performance measures are reported in the Appendix.
Qualitative examples can be found on the project website
\url{http://www.robots.ox.ac.uk/~vgg/research/concealed}.




\begin{table}[] 
  \setlength{\tabcolsep}{2.5pt}
\begin{center}
\begin{tabular}{ l c c c r r r r } 
 \toprule
 & Tr. Occ. & T. Occ.  & Enr.  &  \multicolumn{2}{r}{SDR (dB)} &\multicolumn{2}{r}{WER (\%)} \\  
\addlinespace[2pt]
\backslashbox[20mm]{Method}{\# Spk.}    & & &  & \multicolumn{1}{c}{2} & \multicolumn{1}{c}{3} & \multicolumn{1}{c}{2} & \multicolumn{1}{c}{3}  \\ 
 \midrule
 {GT signal }                            &-&-&-   &   -   &    -   & \multicolumn{2}{c}{20.0}  \\  
 {Mixed signal}                          &-&-&-   &  0.0  & -3.7  & 82.6 & 93.2  \\  
 \midrule
 PIT \cite{yu2016}                       &-&-&            -  &  10.7  &  6.4  & 38.6 & 60.2  \\  
 VoiceFilter \cite{wang2018voicefilter}  &-&-&            pre  &  11.7  &  5.7  & 31.6 & 56.0  \\  
 \midrule
  \multicolumn{8}{c}{No occlusion during evaluation} \\  
  V-Conv \cite{Afouras18}                &\xmark &\xmark &  -    &  12.7 &  9.1  & 24.9 & 33.7  \\  
  V-Conv \cite{Afouras18}                &\cmark &\xmark &  -    &  12.9 &  9.3  & 25.0 & 35.1  \\  
  V-BLSTM                                &\xmark &\xmark &  -    &  12.9 &  9.7  & 24.3 & 33.5  \\  
  V-BLSTM                                &\cmark &\xmark &  -    &  13.0 &  9.5  & 25.3 & 35.7  \\  
  VS                                     &\cmark &\xmark & pre  &  12.8 &  9.2  & 26.5 & 38.3  \\ 
  VS                                     &\cmark &\xmark & self &  12.8 &  9.3  & 26.6 & 40.3  \\ 
 \midrule
   \multicolumn{8}{c}{80\% occlusion during evaluation} \\  
 V-Conv    \cite{Afouras18}              &\xmark &\cmark &  -    &  0.8   & -3.0  & 63.0  & 78.0  \\  
 V-Conv    \cite{Afouras18}              &\cmark &\cmark &  -    &  2.7   & -1.6  & 54.4  & 74.1  \\  

 V-BLSTM                                 &\xmark &\cmark &  -    &  5.8   &  2.4  & 49.3  & 67.1  \\ 
 V-BLSTM                                 &\cmark &\cmark &  -    &  11.6  &  6.3  & 31.3  & 54.3  \\ 
 VS                                      &\cmark &\cmark & pre  &  12.1  &  {\bf 7.3}  & 30.7  & {\bf 50.0}  \\ 
 VS                                      &\cmark &\cmark & self &  {\bf 12.2}  &  7.2  & {\bf 30.3}  & 50.3  \\ 
 \bottomrule

\end{tabular}             
\end{center}
\caption{ 
Evaluation of speech enhancement performance on samples from the LRS3 dataset, for 2 and 3
simultaneous speakers. 
All the samples are 8 seconds long, of which 6 seconds are occluded (3 from either side) when
occlusions are used.  
{\bf Tr. Occ:}  (Train Occlusions) Denotes that the model has been trained using artificial occlusions; 
{\bf T. Occ:} (Test Occlusions) Denotes evaluation with occlusions;
{\bf pre:} Pre-enrollment: the enrollment audio is obtained from a different video of the target speaker;
{\bf self:} Self-enrollment; 
{\bf SDR:} Signal to Distortion Ratio (higher is better);
 {\bf WER:} Word Error Rate from off-the-shelf ASR system (lower is better).
 }
\normalsize        
\label{tab:results}
\vspace{-15pt}
\vspace{-12pt}
\end{table}

\subsection{Baseline models}
We compare our proposed approach to the following baselines and ablations,
which we train and evaluate both with and without visual occlusions. 

\newpara\noindent{\textbf{PIT.}} We implement a blind source separation model that uses only the
noisy audio input stream of Fig. \ref{fig:architecture} and is
trained with a permutation invariant loss following \cite{yu2016}. 
This model is tailored to a predefined number of
speakers.

\newpara\noindent{\textbf{V-Conv.}} This is the convolutional, visually conditioned baseline of
Afouras~{\em et al.}~\cite{Afouras18}. The model uses a series of 1D convolutional blocks for fusing
the audio and video modalities instead of a BLSTM. 
Moreover, the video features are not upsampled by the
video stream as in our proposed model, but the audio-visual fusion is performed at the temporal
resolution of the video frames. The 1D convolutional stack then upsamples the fused input to the
dimensions of the spectrograms.  

\newpara\noindent{\textbf{V-BLSTM.}} This model is similar to our proposed architecture but
conditions only on video features. 


\newpara\noindent{\textbf{VoiceFilter.}} This model conditions only on speaker embeddings and
is equivalent to the sub-network used during the first stage
of the training process. It is essentially a {\em VoiceFilter} \cite{wang2018voicefilter} implementation
with a slightly modified architecture, trained on our dataset.

\newpara\noindent{\textbf{VS.}} Our proposed architecture,
which receives both video and speaker embedding inputs.
As discussed in Section~\ref{sec:method}, we investigate two variants,
{\em VS-pre} and {\em VS-self}, that correspond to the different enrollment methods employed
during evaluation.  


\begin{figure}[t]
  \centering
  \begin{subfigure}{\columnwidth}
      \centering
      \includegraphics[width=\columnwidth]{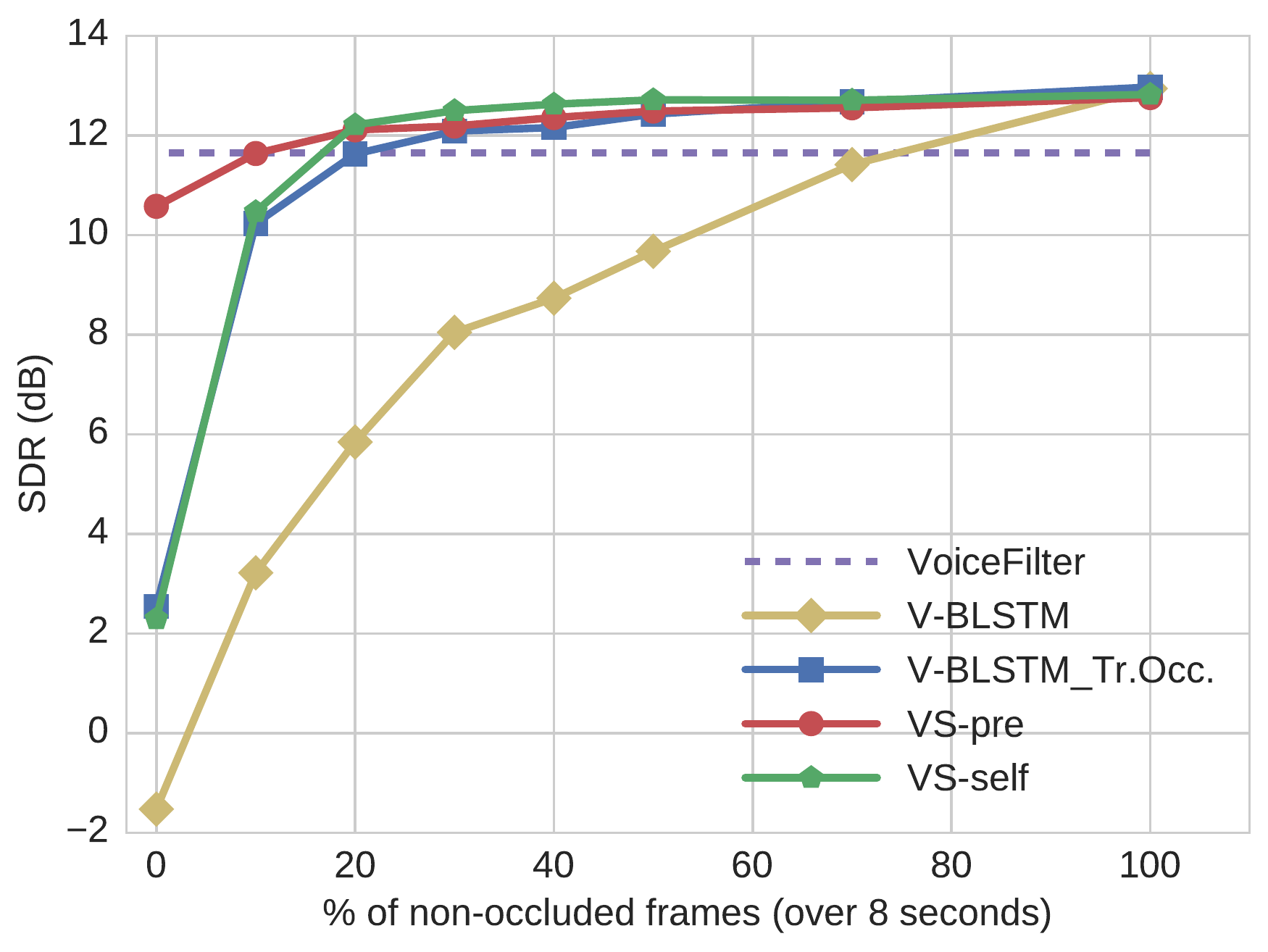}
  \caption{2 Speakers}
  \end{subfigure}
  \begin{subfigure}{\columnwidth}
      \centering
      \includegraphics[width=\columnwidth]{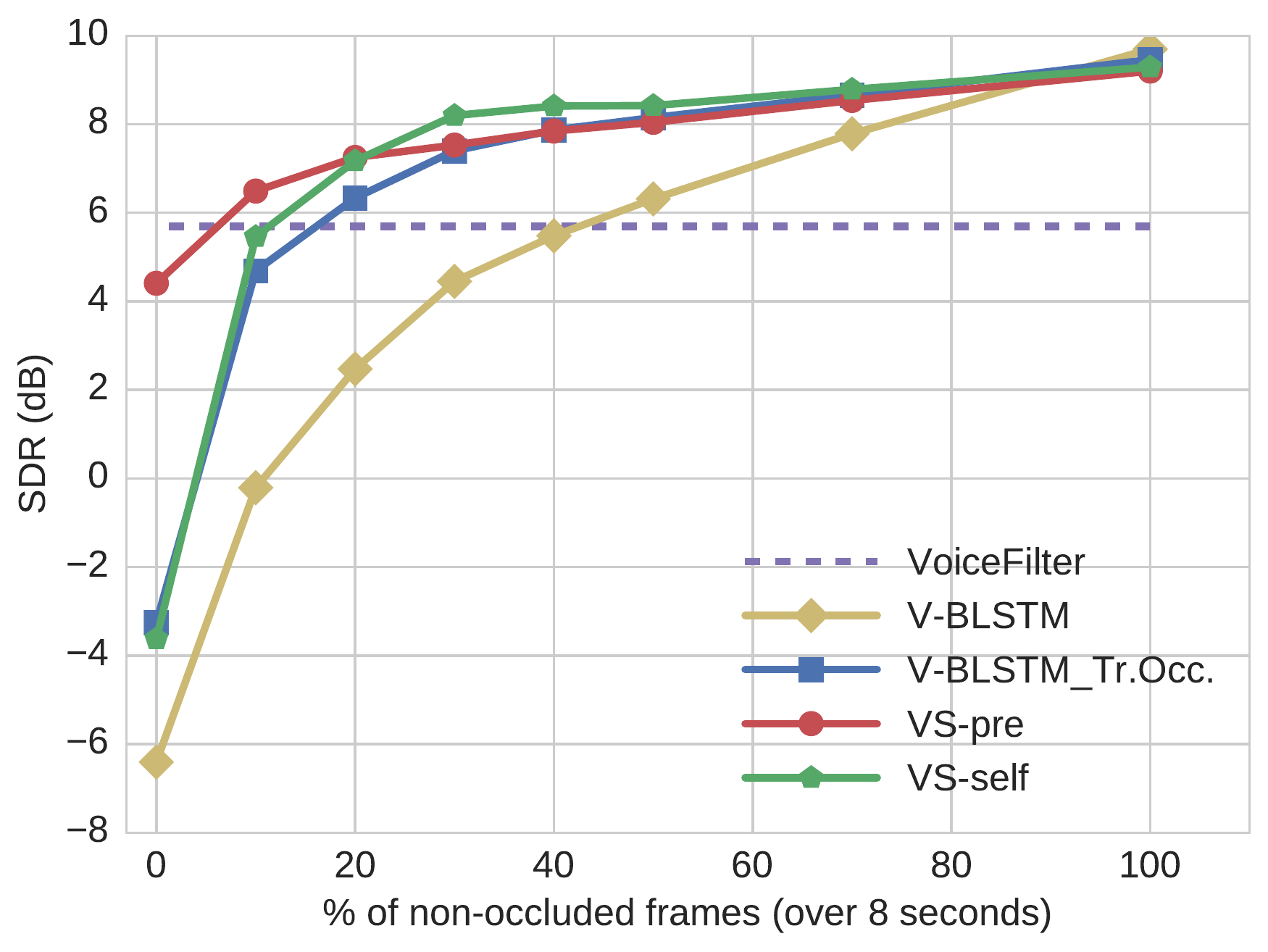}
  \caption{3 Speakers}
  \end{subfigure}
  \caption{ Enhancement performance when occluding varying amounts of the visual input for the 2
Speakers and 3 Speakers scenarios.
    Model notations are explained in the caption of Table \ref{tab:results}.  }
  \vspace{-10pt}
  \label{fig:frame_drop}
\end{figure}
\subsection{Results}

We summarize the results of our experiments in Table~\ref{tab:results}.
When no occlusions are used, the {\em V-BLSTM} model only slightly outperforms {\em V-Conv}.
When $80\%$ of the visual input frames are occluded, the models that haven't
been trained with occlusions fail. Even when we include occlusions during the training of {\em V-Conv},
it cannot deal with the missing visual information, since its receptive field is limited (about 1 second to either
side). On the contrary, {\em V-BLSTM} uses its memory and learns to deal with local occlusions.
Overall however, the proposed {\em VS} models that explicitly condition on the expected speaker
embedding give the best performance.

The results furthermore verify that both the {\em VoiceFilter} and {\em VS-pre} model perform well
when evaluated using enrollment signals from sources different from the target one,
even though they have never been trained in this setting. 


The effect of occluding different amounts of the visual input is studied in Fig.~\ref{fig:frame_drop}.
The {\em V-BLSTM} model that has not been trained on occlusions does not perform well 
when even small parts of the video input are occluded.
When trained with occlusions, {\em V-BLSTM} becomes much more resilient, however it still gives bad
results
for high occlusion percentages and completely fails  when the entire video is occluded. 

The {\em VS-pre} model outperforms {\em V-BLSTM} when half or more of the input is occluded and gives 
similar results for cleaner inputs.

\newpara\noindent{\textbf{Self-enrollment.}} 
For very high occlusion levels,
the initial enhanced estimation of {\em VS-self} is bad and
evidently unable to capture the target voice characteristics.
However, if more than $20\%$ of the frames are clean, self-enrollment performs
best. 
Therefore, apart from the higher occlusion levels, {\em VS} with self-enrollment provides
an advantage compared to {\em V-BLSTM}.
\section{Conclusion}

In this paper, we proposed 
a deep audio-visual speech enhancement
network that is able to separate a speaker's voice by conditioning on
both the speaker's lip movements and/or a representation of their
voice. The network is robust to partial occlusions, and 
the voice representation can be self-enrolled 
from the unoccluded part of the input when it is not possible to obtain
segments for pre-enrollment.
The methods are evaluated on the challenging LRS3 dataset,
and demonstrate performance that exceeds that
of previous state-of-the-art~\cite{Afouras18} when 
the video input is partially occluded.  

\newpara\noindent\textbf{Acknowledgements.}
Funding for this research is provided by the UK EPSRC
CDT in Autonomous Intelligent Machines and Systems, 
the Oxford-Google DeepMind Graduate Scholarship, and the EPSRC 
Programme Grant Seebibyte EP/M013774/1.

\bibliographystyle{IEEEtran}
\bibliography{shortstrings,vgg_other,vgg_local,mybib}

\end{document}